%% file: main.tex
\documentclass[sigconf]{acmart}

\copyrightyear{2026}
\acmYear{2026}
\setcopyright{cc}
\setcctype{by}
\acmConference[CIKM '26]{Proceedings of the 35th ACM International Conference on Information and Knowledge Management}{November 07--11, 2026}{Rome, Italy}
\acmBooktitle{Proceedings of the 35th ACM International Conference on Information and Knowledge Management (CIKM '26), November 07--11, 2026, Rome, Italy}
\acmDOI{10.1145/3799682.3840098}
\acmISBN{979-8-4007-2539-5/2026/11}

\makeatletter
\def\copyrightpermissionfootnoterule{\kern-3\p@\kern2.6\p@}
\makeatother

\usepackage{booktabs}
\usepackage{array}
\usepackage{multirow}
\usepackage{tabularx}
\usepackage{amsmath}

\usepackage{amssymb}
\usepackage{xspace}
\usepackage{microtype}
\usepackage{xcolor}
\usepackage{enumitem}
\usepackage{pifont}
\usepackage{graphicx}
\newcommand{\cmark}{\ding{51}}
\newcommand{\xmark}{\ding{55}}

\newcommand{\system}{HIRA\xspace}
\newcommand{\tauone}{\ensuremath{\tau_{t_1}}\xspace}
\newcommand{\tautwo}{\ensuremath{\tau_{t_2}}\xspace}
\newcommand{\edrm}{EDRM\xspace}

\newcommand{\enumfive}{\texttt{enum5}\xspace}
\newcommand{\macrof}{Macro-F1\xspace}

\title[HIRA: A Human-in-the-Loop Retrieval-Augmented Cascade for Document Classification in Regulated Industries]{%
  \texorpdfstring{\system}{HIRA}: A Human-in-the-Loop Retrieval-Augmented Cascade for Document Classification in Regulated Industries%
}

\author{Shangxuan Tian}
\authornote{Part of this work was conducted while the author was with OCBC.}
\affiliation{%
  \institution{Standard Chartered Bank}
  \city{Guangzhou}
  \country{China}
}
\email{tianshangxuan@u.nus.edu}

\author{Yanhui Chen}
\affiliation{%
  \institution{OCBC}
  \country{Singapore}
}
\email{yanhui79@gmail.com}

\author{Carlos Queiroz}
\affiliation{%
  \institution{OCBC}
  \country{Singapore}
}
\email{caxqueiroz@gmail.com}

\makeatletter
\let\ACM@orig@textbullet\textbullet
\renewcommand{\textbullet}{\ACM@orig@textbullet\nobreak}
\makeatother



\begin{document}

\input{sections/00_abstract}
\maketitle

\input{sections/01_introduction}
\input{sections/02_related_work}
\input{sections/03_problem_setting}
\input{sections/04_system}
\input{sections/05_experimental_setup}
\input{sections/06_results}

\input{sections/08_industrial_validation}
\input{sections/10_limitations}
\input{sections/11_conclusion}

\section*{GenAI Usage Disclosure}
In this work, Generative AI software tools are utilized to edit and improve the quality of existing text, such as spelling, grammar, punctuation, clarity, and engagement, like a typing assistant.

\bibliographystyle{ACM-Reference-Format}
\bibliography{references}

\end{document}

%% file: sections/00_abstract.tex
\begin{abstract}
Document classification in regulated industries is constrained by data
residency, limited cold-start labels, scarce review capacity, and costly
model-governance procedures. In this setting, the practical goal is not
only to obtain high accuracy, but also to reduce LLM calls, avoid frequent
model retraining, and send only the most ambiguous cases to human review.
We present HIRA, a training-free, on-premises retrieval-augmented cascade
for document classification in regulated deployments. HIRA first combines BM25 over OCR
text, dense text embeddings, and image-level representations through
validation-calibrated weighted reciprocal-rank fusion. Confident documents
are classified directly by retrieval. Uncertain or visually confusable
documents are passed to a locally hosted LLM verifier, which receives the
OCR text, retrieved exemplars, label descriptions, and confusion-specific
 terms. When the verifier remains uncertain, the document is
sent to human review. Each correction is then stored as a margin-weighted
retrieval exemplar and also updates a Dirichlet-smoothed confusion graph,
allowing the system to improve without updating model weights.

On a private 80-class trade-finance corpus, {HIRA processes the full
30{,}233-document production stream while requesting human correction for
only 1{,}945 documents (6.4\%)}, improving Macro-F1 from 0.6218 to
\textbf{0.8548}. On the corrected Tobacco-3482 benchmark, HIRA reaches
\textbf{0.9423} Macro-F1 with a locally hosted DeepSeek-R1-Distill-Qwen-32B
verifier, \textbf{17.4 percentage points above} the zero-shot LLM baseline,
while invoking the verifier for only about \textbf{40\%} of documents and
therefore reducing LLM calls by approximately \textbf{60\%}.
With 518 human corrections, corresponding to 24.8\% of the pool, HIRA
matches the fully labelled pool oracle in which all 2{,}086 pool documents
are indexed with their ground-truth labels. These results show that
selective human feedback and retrieval-memory adaptation can provide a
practical alternative to repeated model retraining for long-tail document
classification in regulated deployments.
\end{abstract}

%% file: sections/01_introduction.tex
\section{Introduction}
\label{sec:introduction}

Fine-grained document classification underpins information systems across regulated industries, including financial-services back offices, healthcare records, legal discovery, and government archiving. The literature has largely converged on a single recipe: pre-train a layout-aware backbone, fine-tune on a labelled training set, and deploy. The emergence of large language models (LLMs) and vision--language models (VLMs) has further encouraged a zero-shot default that delegates classification to a frontier API.

Neither approach is straightforward to apply in a regulated deployment. Documents in finance and healthcare cannot leave the deployment environment, ruling out cloud-hosted LLM APIs. At cold start the customer's taxonomy is typically bespoke and the documents are confidential, so a labelled training set for fine-tuning is unavailable. After deployment, document templates drift, new sub-categories appear, and human annotation is the dominant marginal cost. This setting is not hypothetical: the primary motivation for this work is a production trade-finance system classifying 30{,}233 documents across 80 categories, including many near-synonym categories, under strict data-residency requirements.

\begin{figure}[t]
  \centering
  \includegraphics[width=1.0\linewidth, trim=0 28mm 0 25mm, clip=true]{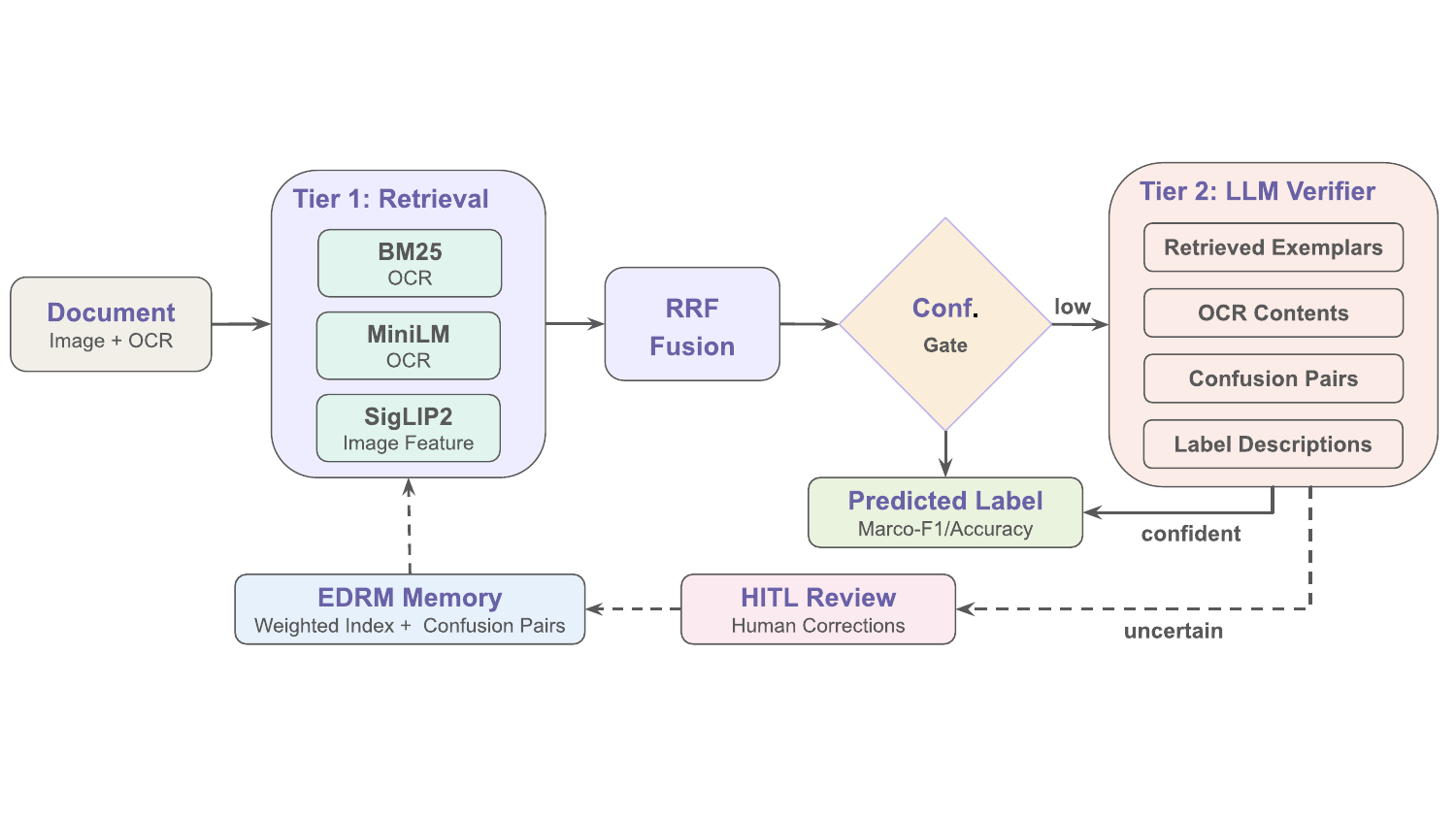}
  \caption{The proposed \system{} cascade framework. }
  \label{fig:cascade}
\end{figure}

We propose \textbf{\system{}}, a training-free retrieval-augmented classification cascade designed for this regime. The pipeline is shown in Figure~\ref{fig:cascade}. \system{} fuses three retrieval signals (BM25 over OCR text, dense sentence embeddings, and image-level representations) using validation-calibrated weighted reciprocal-rank fusion, and routes only uncertain or visually confusable cases to a verifier LLM. The verifier receives the OCR text, the top retrieved exemplars, label descriptions, and distinguishing terms mined from past misclassifications, and produces a structured prediction with an ordinal confidence bucket. Low-confidence verifications enter a human-in-the-loop (HITL) queue, and each correction updates the system in two ways: as a margin-weighted entry in the retrieval index, and as an update to a Dirichlet-smoothed confusion graph that augments later verification prompts. No model parameters are updated at any point; only the retrieval memory and a lightweight confusion graph are mutable.

On a private 80-class trade-finance corpus (Financial-80), \system{} processes the full 30{,}233-document production stream and requests human correction for only 1{,}945 documents (6.4\%), improving \macrof{} from 0.6218 to 0.8548. On the corrected Tobacco-3482 benchmark~\cite{kumar2012learning, lim2024labelerrors}, \system{} obtains \macrof{} 0.9423 with a locally hosted DeepSeek-R1-Distill-Qwen-32B verifier, 17.4 percentage points above the LLM zero-shot baseline at a 40\% verifier call rate, and matches the fully labelled pool oracle (Section~\ref{sec:h2}) after labelling 24.8\% of the incoming stream.

\paragraph{Contributions.} The paper makes three contributions:

\begin{itemize}[leftmargin=*,itemsep=2pt,topsep=2pt]
  \item Cold-start adaptation under a small annotation budget. On Financial-80 (80 categories, 30{,}233 production documents), \system{} reaches \macrof{} 0.8548 with only 1{,}945 corrections (6.4\% of the stream), and the queue rate falls to 6.2\% as the index grows. On Tobacco-3482, 24.8\% of the pool labelled is enough to match the fully labelled oracle.
  \item A training-free retrieval-augmented generation (RAG) cascade that exceeds LLM zero-shot. \system{} reaches \macrof{} 0.9423 on corrected Tobacco-3482, $+$17.4 points over a locally hosted DeepSeek-R1-Distill-Qwen-32B zero-shot baseline, at a verifier call rate of approximately 40\%.
  \item \edrm{} (Evolving Document Relevance Memory), a dual-signal HITL memory. Each correction updates a margin-weighted retrieval index and a Dirichlet-smoothed confusion graph; the graph feeds distinguishing terms into the Tier-2 verifier prompt for confusion pairs.
\end{itemize}

%% file: sections/02_related_work.tex
\section{Related Work}
\label{sec:related}

Existing research on document classification can be classified into two categories: (i) fine-tuned document-image classification methods, and (ii) retrieval-augmented and cascade inference approaches.

\paragraph{Fine-tuned document-image classification.}
Layout-aware and OCR-free document models such as LayoutLM~\cite{xu2020layoutlm}, LayoutLMv2~\cite{xu2021layoutlmv2}, LayoutLMv3~\cite{huang2022layoutlmv3}, DiT~\cite{li2022dit}, Donut~\cite{kim2022donut}, and UDOP~\cite{tang2023udop} have become standard supervised baselines for document understanding. Two structural limitations apply to our setting: (i) they require a labelled training set, unavailable at cold start, and (ii) adapting to taxonomy changes requires a re-fine-tuning cycle that adds a GPU pass and a compliance review. \system{} differs in regime --- it does not update weights, only the retrieval index --- which avoids both.

\paragraph{Retrieval-augmented classification.}
Retrieval-Augmented Generation~\cite{lewis2020rag} and dense passage retrieval~\cite{karpukhin2020dpr} showed that a non-parametric retrieval component can supplement learned parameters for knowledge-intensive tasks. For classification, $k$NN-LM~\cite{khandelwal2020knnlm} aggregates labels from nearest neighbours in a frozen-corpus index, $k$NN-Prompting~\cite{xu2023knnprompting} extends this to LLM in-context learning, Class-RAG~\cite{chen2024classrag} couples retrieval with an LLM verifier for content-moderation classification, and SRA~\cite{mao2025sra} applies retrieval-augmentation to long-tail legal text classification. Our \edrm{} updates this index online from each correction: a margin-weighted exemplar plus a Dirichlet-smoothed confusion graph that feeds distinguishing terms back into the verifier's prompt for known confusion pairs.

\paragraph{Cascade inference and cost-aware routing.}
Cascading models of increasing capacity to balance cost and quality is a classical technique~\cite{dohan2022cascades}. FrugalGPT~\cite{chen2023frugalgpt} frames cascade routing as an optimisation problem and reduces LLM calls by deferring only uncertain queries to a stronger model. CALM~\cite{schuster2022calm} achieves similar cost reductions via early-exit confidence signals within a single model. These approaches assume either no retrieval or a fixed retrieval stage. In \system{}, the Tier-1 and Tier-2 thresholds are calibrated on a validation set, but the underlying retrieval index evolves as corrections are incorporated.

%% file: sections/03_problem_setting.tex
\section{Problem Setting}
\label{sec:problem}

Let $\mathcal{X}$ denote the document space and $\mathcal{Y}=\{1,\ldots,C\}$ a customer-specific label taxonomy. Given a document $x\in\mathcal{X}$ (scanned or digital PDF) in a regulated deployment --- financial services, healthcare, legal discovery --- the goal is to predict a class $y\in\mathcal{Y}$. Four constraints jointly preclude both supervised fine-tuning and cloud-LLM defaults:

\begin{itemize}[label={},leftmargin=0pt,itemsep=1pt,topsep=2pt]
  \item \textbf{(1) Cold start under drift.} At deployment, only tens of labelled documents are available per class. Each customer has a custom taxonomy, and the data are long-tailed. As document templates change, a fixed classifier can gradually lose accuracy without being immediately noticed.
  \item \textbf{(2) Residency.} Documents must not leave the deployment environment, excluding cloud LLM APIs as a runtime dependency. This constraint follows from data-residency and bank-secrecy regulations common to financial services and healthcare, and it applies to every stage of the pipeline, not only the final classifier: OCR, embedding, and verification must all run on infrastructure the customer controls.
  \item \textbf{(3) Compliance-bound retraining.} In a regulated bank, retraining and redeploying a model triggers a model-governance review that adds significant process latency and nontrivial cost. Weight-update-based adaptation is rarely permissible at the cadence that drift requires. A governance review typically involves revalidating the model against a held-out compliance suite and obtaining sign-off from risk and legal stakeholders, a cycle that can take weeks even for a minor weight update.
  \item \textbf{(4) Audit trail and bounded annotation budget.} Every decision must be explainable as a reference to similar documents already on file, which favours retrieval-style architectures over opaque end-to-end classifiers whose predictions cannot be traced back to a concrete precedent. HITL review is the slowest and most expensive step in the pipeline, and review capacity is fixed regardless of document volume; the appropriate metric is therefore accuracy per human label rather than raw accuracy.
\end{itemize}

Retrieval-augmented classification fits naturally: only the retrieval memory and a lightweight confusion graph are updated from HITL corrections, without touching model weights.

%% file: sections/04_system.tex
\section{The HIRA System}
\label{sec:system}

\system{} is a two-stage retrieval-augmented cascade with an HITL feedback loop. Figure~\ref{fig:cascade} shows the architecture. All components are training-free post-deployment; the only state that changes over a deployment's lifetime is the retrieval index and a Dirichlet-smoothed confusion graph, both maintained by the Evolving Document Relevance Memory (\edrm) described in Section~\ref{sec:edrm}.

Figure~\ref{fig:cascade} traces one document through the cascade: Tier~1 (Section~\ref{sec:tier1}) fuses three retrievers via weighted RRF and short-circuits when the top class clears $\tauone$; otherwise Tier~2 (Section~\ref{sec:tier2}) escalates to a locally hosted LLM verifier, which accepts a label when its confidence clears $\tautwo$ or else routes the document to the HITL queue. Every human correction feeds back into \edrm{} (Section~\ref{sec:edrm}), updating the retrieval index and confusion graph that inform future decisions, without ever updating model weights.

\subsection{Tier 1: Multi-Modal Retrieval with RRF Fusion}
\label{sec:tier1}

Tier 1 runs three retrievers in parallel over the current index, each returning a top-$k$ list:
\begin{itemize}[leftmargin=*,itemsep=2pt,topsep=2pt]
  \item \textbf{BM25} over raw OCR text (PaddleOCR~\cite{li2022paddleocr}).
  \item \textbf{MiniLM} embeddings over OCR (all-MiniLM-L6-v2~\cite{wang2020minilm}).
  \item \textbf{SigLIP2} image embeddings~\cite{tschannen2025siglip2}.
\end{itemize}

The three ranked lists are fused via weighted reciprocal-rank fusion (RRF)~\cite{cormack2009rrf}:
\begin{equation}
\mathrm{score}(d) = \sum_{r \in \mathcal{R}} w_r \cdot \frac{1}{k_{\mathrm{rrf}} + \mathrm{rank}_r(d)}
\label{eq:rrf}
\end{equation}
where $\mathcal{R} = \{\mathrm{BM25}, \mathrm{MiniLM}, \mathrm{SigLIP2}\}$ is the set of retrievers, $k_{\mathrm{rrf}} = 10$ is the rank constant that dampens the contribution of lower-ranked results, and $w_r$ are per-retriever weights calibrated per deployment by sweeping single-feature \macrof{} on the validation split. RRF fusion improves on any single retriever, as reported in Section~\ref{sec:results}.

\paragraph{Per-class aggregation and the Tier-1 gate.}
Per-document scores are aggregated into per-class scores, with \edrm{} entry weights modulating each indexed exemplar's contribution. For class $c$ with indexed exemplars $\mathcal{I}_c$,
\begin{equation}
S(c) = \sum_{d \in \mathcal{I}_c} w(d) \cdot \mathrm{score}(d), \qquad s(c) = \frac{S(c)}{\sum_{c'} S(c')}
\label{eq:perclass}
\end{equation}
where $w(d)$ is the \edrm{} entry weight (Equation~\ref{eq:margin-weight}, $w(d)=1$ for seed exemplars never corrected). Let $\hat{c}_1, \hat{c}_2$ denote the top-two classes ordered by $s(c)$; $\hat{c}_1$ is the top prediction.

Tier 1 accepts $\hat{c}_1$ when $s(\hat{c}_1) \geq \tauone$ (Section~\ref{sec:lesson-tau1}); the cascade then short-circuits and returns $\hat{c}_1$. Cases that do not satisfy this condition are escalated to Tier 2, which supplies the LLM with additional context for verification.

\subsection{Tier 2: LLM Contrastive Verification}
\label{sec:tier2}

Tier 2 invokes a verifier LLM on the document's OCR text, Tier-1's top-$k$ retrieved exemplars $\mathcal{E}_1(d)$ ($k{=}3$ for Tobacco-3482, $k{=}5$ for Financial-80), and the category-description block. Let $\mathcal{L}_1(d)$ denote the set of labels appearing in $\mathcal{E}_1(d)$. When $(\hat{c}_1, \hat{c}_2)$ forms a high-confusion pair in $\mathcal{H}$ (Section~\ref{sec:edrm}), the prompt additionally includes distinguishing terms mined from prior corrections of that pair. The deployed verifier is a locally hosted DeepSeek-R1-Distill-Qwen-32B~\cite{deepseekai2025r1}, constrained via guided generation to output a predicted label, an ordinal confidence bucket, and a free-text justification (Section~\ref{sec:enum5-fix}).

\paragraph{The Tier-2 gate.}
Let $\tilde{c}$ and $\tilde{\mathrm{conf}}$ denote the verifier's outputs. Tier 2 accepts when
\begin{equation}
\tilde{c} \in \mathcal{L}_1(d) \quad \mathrm{and}\quad \tilde{\mathrm{conf}} \geq \tautwo,
\label{eq:tier2gate}
\end{equation}
where \tautwo{} is the verifier accept threshold (Section~\ref{sec:lesson-tau2}). Cases that fail either condition are routed to the HITL queue.

\subsection{EDRM: Evolving Document Relevance Memory}
\label{sec:edrm}

Each HITL correction triggers two updates to \edrm.

\paragraph{(a) Time-decayed difficulty-aware exemplar insertion.}
For each correction $(d, c_\mathrm{true})$, the true-class margin
\begin{equation}
m_{\mathrm{true}} = s(c_\mathrm{true}) - \max_{c \neq c_\mathrm{true}} s(c)
\label{eq:true-margin}
\end{equation}
measures how confidently Tier-1 ranked the correct class (negative when Tier-1 misclassified $d$). The correction is inserted into the retrieval index with weight
\begin{equation}
w(d) = \mathrm{clip}\!\left(\exp\!\left(-\,m_{\mathrm{true}} / \gamma\right),\; w_{\min},\; w_{\max}\right),
\label{eq:margin-weight}
\end{equation}
($\gamma = 0.1$, $w_{\min}=0.1$, $w_{\max}=10$), so misclassified and low-margin corrections receive larger weights than easy confirmations; clipping prevents a single confidently wrong correction from dominating the retrieval vote. Index entries decay multiplicatively as corrections accumulate; entries below $w_{\min}$ after decay are pruned. The decay time constant $\tau_{\mathrm{index}}$ is a deployment parameter set against the customer's drift profile (Section~\ref{sec:lesson-decay}).

\paragraph{(b) Dirichlet-smoothed confusion graph.}
\edrm{} also maintains a $K{\times}K$ confusion-count matrix $C$, where $C_{ij}$ counts predictions of class $i$ for true-label $j$. The Dirichlet-smoothed posterior
\begin{equation}
\hat{P}(c_\mathrm{true}=j \mid \hat{c}_1=i) = \frac{C_{ij} + \alpha}{\sum_{k} C_{ik} + \alpha K}
\end{equation}
($\alpha = 0.1$) identifies high-confusion pairs (posterior ${>}\,\theta_{\mathrm{conf}} = 0.15$), added to $\mathcal{H}$. For each such pair, the top distinguishing OCR tokens from past corrections are stored and injected into the Tier-2 prompt. Graph counts decay with half-life $\tau_{\mathrm{graph}}$ to discard outdated confusion patterns.

\subsection{HITL Loop and Frozen-State Evaluation}
\label{sec:hitl}

The protocol separates three concerns often conflated in HITL evaluation. (i)~\emph{Shared stream order}: all compared methods process the same documents in the same fixed order and mini-batch size. (ii)~\emph{Per-method HITL selection}: each method decides independently which documents require a human label, using its own uncertainty gate, so the methods consume the same arrival stream but build different correction sequences --- precisely what the annotation-cost comparison is intended to measure. (iii)~\emph{Shared frozen-state evaluation}: at each checkpoint of $N$ corrections the resulting \edrm{}-augmented index is frozen and the full cascade (Tier 1 plus the real-LLM Tier 2) is re-run on the same held-out test set across methods.

%% file: sections/05_experimental_setup.tex
\section{Experimental Setup}
\label{sec:setup}

\subsection{Datasets}
\label{sec:datasets}

\paragraph{Financial-80, primary deployment corpus.}
A private 80-category trade-finance document corpus drawn from a production financial-services pipeline. The distribution is strongly skewed and a large fraction of categories are near-synonyms that require fine-grained disambiguation (e.g., multiple invoice and certificate subtypes). Under the data-sharing agreement we report only aggregate metrics (overall \macrof{}, head/medium/tail group F1, verifier call rate, HITL queue rate); per-class F1 tables and category names cannot be released. The class-count distribution, anonymised as C1\ldots C80, is shown in Section~\ref{sec:industrial}.

\paragraph{Tobacco-3482 (corrected), public benchmark.}
The corrected Tobacco-3482 subset~\cite{lim2024labelerrors} re-annotates the original corpus~\cite{kumar2012learning} and removes 21\% of multi-label or empty rows, yielding 2{,}737 documents (2{,}186 train / 272 val / 279 test). For the HITL-curve experiments, 100 seed documents (10 per class) initialise the index, 2{,}086 pool documents are streamed in a fixed random order, and 279 test documents are held out. Tier-1 RRF weights and \tauone{} are calibrated on the validation split. The ten categories (Email, Letter, Memo, Form, Report, Scientific, ADVE, Note, News, Resume) exhibit a roughly $5\times$ head-to-tail ratio. OCR is produced with PaddleOCR~\cite{li2022paddleocr}.

\subsection{Baselines}
\label{sec:baselines}

All baselines share \system{}'s retrieval index, pool, and test split.

\begin{itemize}[leftmargin=*,itemsep=2pt,topsep=2pt]
  \item \textsf{bm25\_only}, \textsf{minilm\_only}, \textsf{siglip2\_only}: single-feature retrieval with top-1 voting.
  \item \textsf{RRF fusion}: Reciprocal Rank Fusion of the three retrievers above.
  \item \textsf{kNN + Naive HITL}: RRF with corrections injected at uniform weight $1.0$, without \edrm{}.
  \item \textsf{LLM zero-shot (DeepSeek)}: locally hosted DeepSeek-R1-Distill-Qwen-32B, prompted on OCR text with our system prompt and guided\_json output.
  \item \textsf{DiT fine-tuned (seed / HITL-augmented)}: Facebook Document Image Transformer~\cite{li2022dit} fine-tuned on the 100 seed documents (\textsf{DiT-100}), and separately on the same 100 seed documents plus the 518 \system{}-selected corrections (\textsf{DiT-618}, equal annotation budget to \system{} at $N{=}518$).
\end{itemize}

\begin{table}[!ht]
\centering
\caption{Static baselines on corrected Tobacco-3482 (seed-only index, 10 docs/class). \system{} at HITL convergence (N=518) is reported in Table~\ref{tab:hitl-trajectory}.}
\label{tab:block0}
\resizebox{\linewidth}{!}{%
\begin{tabular}{lrrrrr}
\toprule
Method & \macrof{} & Acc. & Head-F1 & Med-F1 & Tail-F1 \\
\midrule
BM25 only          & 0.4758 & 0.5161 & 0.623 & 0.365 & 0.477 \\
MiniLM only        & 0.4418 & 0.4588 & 0.518 & 0.333 & 0.511 \\
SigLIP2 only       & 0.7718 & 0.7670 & 0.731 & 0.789 & 0.790 \\
{RRF fusion}       & 0.7736 & 0.7778 & 0.778 & 0.798 & 0.737 \\
\midrule
{DS 0-shot}        & 0.7683 & 0.8172 & 0.928 & 0.730 & 0.660 \\
DiT-100\textsuperscript{†}  & 0.8194 & 0.8065 & 0.839 & 0.841 & 0.971 \\
\bottomrule
\end{tabular}%
}
\par\smallskip\footnotesize\raggedright
\textsuperscript{†}DiT fine-tuned on the same 100 seed documents.
\end{table}

\paragraph{LLM verifier selection.}
We evaluated three locally hosted LLMs as Tier-2 verifier candidates on corrected Tobacco-3482 at $N{=}0$ (seed-only index): DeepSeek-R1-Distill-Qwen-32B~\cite{deepseekai2025r1} (\macrof{} 0.8819), Qwen-VL2.5~\cite{bai2025qwen25vl} (\macrof{} 0.8452), and Llama-3.1-8B~\cite{grattafiori2024llama3} (\macrof{} 0.8233). DeepSeek outperformed both alternatives and was also faster per document on the same A100, so we use it as the deployed verifier in all remaining experiments.

\paragraph{Supervised fine-tuning scope.}
We include DiT fine-tuned at two label counts (\textsf{DiT-100} and \textsf{DiT-618}) matched to \system{}'s seed-only and $N{=}518$ checkpoints. Fully supervised fine-tuning on a training-set-sized corpus (e.g., LayoutLMv3 on the full training split) requires a labelled training set unavailable at cold start and triggers a model-governance review on each update, so we do not include it as a primary baseline.

\subsection{Metrics and Protocol}
\label{sec:metrics}

The primary metric is \macrof{}, which weights all categories equally and is therefore appropriate for the imbalanced setting. We additionally report accuracy, head/medium/tail F1, HITL queue rate, verifier call rate, and Tier-1/Tier-2 invocation fractions as deployment-cost diagnostics. All HITL-curve numbers use the frozen-state, full-cascade protocol of Section~\ref{sec:hitl}.

\paragraph{Variance and seed protocol.} \system{}'s stochasticity is confined to the Tier-2 verifier. A three-run check on Financial-80 at $N{=}1{,}945$ yields \macrof{} = 0.8523 / 0.8535 / 0.8586 (mean 0.8548, range 0.63 points), an order of magnitude below the smallest method gap; given this and the compute cost of full HITL trajectories, all main results use a single fixed seed.

%% file: sections/06_results.tex
\section{Results}
\label{sec:results}

We report results across two corpora. Section~\ref{sec:fin80} reports production deployment results on Financial-80 (80 categories, 30{,}233 documents); the present section reports controlled benchmark results on Tobacco-3482 (10 categories, 2{,}086 pool documents), which calibrate the V3 configuration transferred to Financial-80.

\subsection{Static baselines (no HITL)}
\label{sec:block0}

Table~\ref{tab:block0} reports baseline results on the corrected Tobacco-3482 split with a seed-only retrieval index (10 documents per class, 100 in total). With only 100 seed documents, RRF fusion (0.7736) already matches the DeepSeek zero-shot baseline (0.7683). DS zero-shot's 26.8-point Head--Tail F1 gap (versus 4.1 for RRF) shows why \macrof{}, not accuracy, is the appropriate metric here. DiT fine-tuned on the same 100 seed documents reaches 0.8194 \macrof, ahead of the retrieval-only baselines but $-6.3$ points behind the \system{} cascade at $N{=}0$ (0.8819).

\subsection{\system{} cascade vs.\ LLM zero-shot}
\label{sec:h1}

Table~\ref{tab:h1} summarises the cascade-versus-zero-shot comparison. With a locally hosted DeepSeek-R1-Distill-Qwen-32B verifier under the \enumfive{} configuration (Section~\ref{sec:enum5-fix}), \system{} reaches \macrof{} 0.9423 after HITL convergence on a 2{,}086-document pool. This is $+17.4$ points above the DeepSeek zero-shot baseline (0.7683), at a verifier call rate of approximately 40\% versus 100\% for zero-shot. RRF fusion alone, with the 100-document seed index, already matches LLM zero-shot ($+0.53$ points); HITL enrichment and \edrm{} account for the remainder of the $+17.4$-point gain.

\begin{table}[t]
\centering
\small
\caption{\system{} cascade vs.\ LLM zero-shot on corrected Tobacco-3482.}
\label{tab:h1}
\begin{tabular}{lrr}
\toprule
System & \macrof{} & Verifier call rate \\
\midrule
DS 0-shot & 0.7683 & 100\% \\
RRF fusion                        & 0.7736 & 0\% \\
\textbf{\system{}} & \underline{0.9423} & $\sim$40\% \\
\midrule
$\Delta$(\system{} $-$ DS 0-shot) & \underline{$+$17.4 pp} & $-60$\% \\
\bottomrule
\end{tabular}
\end{table}

\subsection{HITL gains and label cost}
\label{sec:h2}

\begin{figure}[t]  
  \centering
  \includegraphics[width=1.0\linewidth]{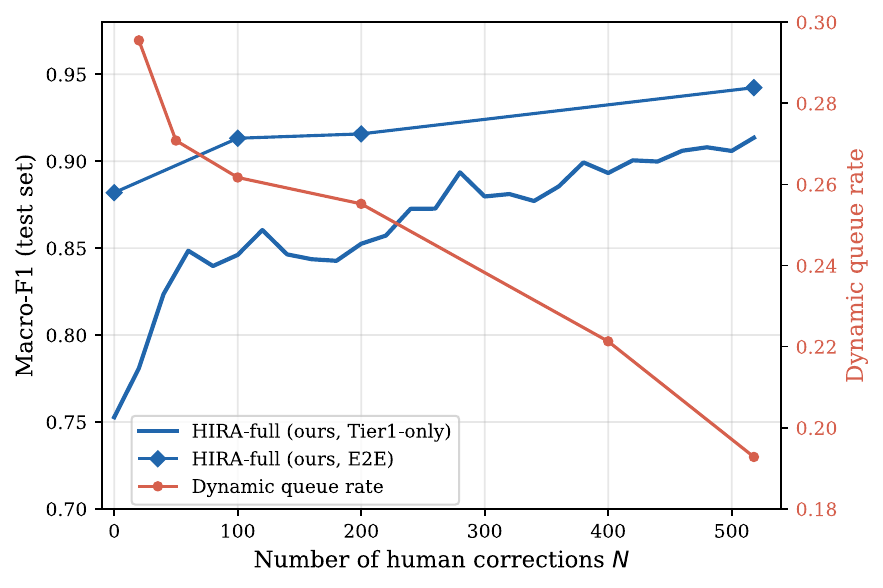}
  \caption{\system{} HITL improvement curve on corrected Tobacco-3482 (primary axis, green). Secondary axis (red): HITL dynamic queue rate.}
  \label{fig:hitl-curve}
\end{figure}

Figure~\ref{fig:hitl-curve} plots the test-set \macrof{} of the frozen-state full cascade as a function of the number of HITL corrections $N$, under the default configuration ($\tauone=0.675$, $\tautwo=0.95$, $\tau_{\mathrm{index}}=10^4$). Table~\ref{tab:hitl-trajectory} reports the same trajectory in tabular form, and Table~\ref{tab:hitl-configs} compares alternative configurations.

\paragraph{Annotation savings.}
At $N{=}518$, \system{} has labelled 24.8\% of the 2{,}086-document incoming stream and reaches \macrof{} 0.9423, matching the fully labelled pool oracle, in which every pool document is inserted into the retrieval index with its ground-truth label at uniform weight $1.0$. In other words, labelling less than a quarter of the incoming stream is sufficient to recover oracle performance.

\paragraph{Comparison with limited-label supervised fine-tuning.}
At equal annotation budget (618 labels total), DiT fine-tuned on the same documents reaches \macrof{} 0.9193, compared with \system{}'s 0.9423 --- a $+2.3$-point advantage with no weight update, training loop, or checkpoint management. DiT achieves higher tail-F1 (0.978 vs.\ 0.970), consistent with its visual pre-training aligning well with the distinctive layouts of tail categories (Note, News, Resume). \system{} leads on medium classes (0.926 vs.\ 0.915), where the confusion-graph mechanism provides gains that visual features alone cannot capture.

\begin{table}[t]
\centering
\small
\caption{EDRM contribution ablation on corrected Tobacco-3482 (Tier-1 only).}
\label{tab:edrm-ablation}
\resizebox{\linewidth}{!}{%
\begin{tabular}{lccrrrrr}
\toprule
System & EDRM  & \macrof{} & Acc. & Head & Med & Tail \\
\midrule
kNN+Naive HITL  & \xmark  & 0.8857 	& 0.8823 	& 0.875 	& 0.854 	& 0.936 \\
\system{} & \cmark  & 0.9135	& 0.8961	 & 0.871	& 0.907	& 0.964 \\
\bottomrule
\end{tabular}%
}
\end{table}

\paragraph{EDRM versus naive correction injection.}
Table~\ref{tab:edrm-ablation} isolates the \edrm{} contribution at Tier-1 ($N{=}518$, no Tier-2 verifier). Margin-weighted exemplar injection raises \macrof{} from 0.8857 (naive, uniform weight) to 0.9135 ($+2.78$ points), with gains concentrated in medium classes (0.854$\to$0.907). When the Tier-2 verifier is added, the full \system{} cascade reaches 0.9423 versus \textsf{kNN+Naive HITL}'s 0.9135 ($+2.88$ points): the confusion graph populated by \edrm{} augments the verifier's distinguishing-term prompt, so the retrieval-level gain propagates and compounds through the verification stage.

\paragraph{Configuration ablation.}
Table~\ref{tab:hitl-configs} compares four operating points: \textsf{V0} (aggressive decay, $\tau_{\mathrm{index}}=300$), \textsf{V1} (low decay, $\tautwo=0.9$), \textsf{V2} (raises \tauone{} to 0.85), and \textsf{V3} (the deployed configuration). \textsf{V3} achieves the highest cascade-final \macrof{} (0.9423) with 40\% fewer corrections than \textsf{V0} (518 versus 862).

\paragraph{Cascade efficiency.}
At the deployed V3 operating point, about 60\% of test documents are short-circuited at Tier~1 and about 40\% reach Tier~2. The dynamic queue rate, namely the fraction of the incoming document stream that ultimately enters the HITL queue, is 18.6\% at the V3 checkpoint; the remaining cases are accepted by the verifier without human review. Figure~\ref{fig:hitl-curve} (secondary axis) shows that the queue rate decreases as the index matures, a self-reinforcing dynamic also observed on Financial-80 (Section~\ref{sec:fin80}). The verifier's net contribution shrinks correspondingly ($+9.67$ under V0, $+2.88$ under V3 in Table~\ref{tab:hitl-configs}), so LLM dependence decreases with deployment age.

\begin{table}[t]
\centering
\small
\caption{HITL learning trajectory with end-to-end evaluation on corrected Tobacco-3482.}
\label{tab:hitl-trajectory}
\begin{tabular}{lrrrrr}
\toprule
$N$ & \macrof & Acc.\ & Head & Med & Tail \\
\midrule
DS 0-shot & 0.7683 & 0.8172 & 0.928 & 0.730 & 0.660 \\
\midrule
0   & 0.8819 & 0.8925 & 0.935 & 0.859 & 0.859 \\
100 & 0.9132 & 0.9176 & 0.952 & 0.875 & 0.953 \\
200 & 0.9157 & 0.9211 & 0.936 & 0.892 & 0.948 \\
518 & \underline{0.9423} & 0.9355 & 0.937 & 0.926 & \underline{0.970} \\
DiT-618\textsuperscript{†} & 0.9193 & 0.9211 & 0.956 & 0.915 & 0.978 \\
\midrule
Oracle-2086\textsuperscript{*} & \underline{0.9423} & \underline{0.9462} & \underline{0.957} & \underline{0.930} & 0.944 \\
\bottomrule
\end{tabular}
\par\smallskip\footnotesize\raggedright
\textsuperscript{*}Oracle-2086 denotes a fully labelled retrieval oracle in which all 2{,}086 pool documents with ground-truth labels are inserted into the retrieval index with uniform weight $1.0$. \\
\textsuperscript{†}DiT fine-tuned on 100 seed + 518 \system{}-selected corrections.
\end{table}

\begin{table}[t]
\centering
\small
\caption{Four-way HITL configuration comparison on corrected Tobacco-3482.}
\label{tab:hitl-configs}
\resizebox{\linewidth}{!}{%
\begin{tabular}{lrrrrr}
\toprule
Config ($\tau_{\mathrm{gate}}$, $\tau_{\mathrm{index}}$) & Corr. & Ann. rate & T1-only & C-peak & C-final \\
\midrule
V0: aggressive decay (0.9, 300)         & 862 & 41\%   & 0.8337 & 0.9479 & 0.9304 \\
V1: lower Tier-2 gate (0.9, 10k)        & 265 & 13\%   & 0.8689 & 0.9248 & 0.9069\\
V2: tighter Tier-1 gate (0.85, 10k)     & 298 & 14.3\% & 0.8667 & 0.9080 & 0.9056\\
\textbf{V3: deployed (0.95, 10k)}       & \textbf{518} & \underline{24.8\%} & \underline{0.9135} & \underline{0.9423} & \underline{0.9423} \\
\bottomrule
\end{tabular}%
}
\par\smallskip\footnotesize\raggedright
``T1-only'': retrieval ceiling without verifier.
``C-final'': frozen-state full cascade at the listed $N$.
``C-peak'': maximum \macrof{} across the trajectory (peak $N$: 400/200/200/518 for V0--V3).
``Ann. rate'': fraction of pool corrected (corrections/2{,}086).
$\tau_{\mathrm{gate}}$: V0, V1, V3 use $\tautwo{}$; V2 uses $\tauone{}$ ($\tautwo=0.9$).
\end{table}

\subsection{Cascade configuration: three key parameters}
\label{sec:h4}
\label{sec:lessons}
\label{sec:lesson-tau1}
\label{sec:lesson-tau2}
\label{sec:lesson-decay}
\label{sec:enum5-fix}

Three configuration parameters govern the quality-cost trade-off of the cascade, and each carries a non-obvious lesson.

\paragraph{\tauone{}: calibrate on validation rather than from a prior.}
The Tier-1 accept threshold was calibrated by sweeping $\tauone \in [0.50, 0.90]$ on the corrected validation split and selecting the Pareto plateau on (\macrof, Tier-2 invocation rate); we obtain $\tauone=0.675$. Above $\tauone=0.85$ a false cliff appears: over-routing to Tier~2 incurs LLM cost without quality gain, because the verifier rarely overturns a borderline-confident Tier-1 prediction.

\paragraph{\tautwo{} rather than \tauone{} is the queue-rate lever.}
Comparing V1 and V2 shows that tightening the Tier-1 gate from the default setting changes the annotation rate only marginally (13\% to 14.3\%) and reduces cascade-peak \macrof{}. Raising $\tautwo$ from $0.9$ to $0.95$ (V1$\to$V3) increases the queue rate to 18.6\% and improves cascade-final \macrof{} by $3.54$ points: in a dense-index regime, it is the Tier-2 accept threshold rather than the Tier-1 gate that controls human-review volume. One practical prerequisite is that small open-weight LLMs collapse numeric confidence to the visible gate value (0.950); replacing it with an ordinal \enumfive{} enum~\cite{willard2023outlines} restores a usable spread, and the fix applies to any small-LLM cascade.

\paragraph{Index decay $\tau_{\mathrm{index}}$ is deployment-specific.}
On a static benchmark a long decay ($\tau_{\mathrm{index}}=10^4$, config V3) dominates: the index is denser, Tier-1 short-circuits more often, and fewer corrections are needed. In production, where documents genuinely drift, a finite decay is necessary in order to flush stale exemplars. There is no universal best value; operators should set $\tau_{\mathrm{index}}$ from the drift profile of their deployment. An alternative is calendar-time decay: entries older than a fixed period (e.g.\ six months) are expired regardless of correction volume, which is preferable when document templates change on a known schedule such as regulatory form revisions.

\subsection{On-premises deployment configuration}
\label{sec:h3}

The result in Table~\ref{tab:h1} is obtained with a locally hosted DeepSeek-R1-Distill-Qwen-32B verifier served by vLLM~\cite{kwon2023vllm} on a single A100, accessed over an internal network. No document, retrieval snippet, or prompt leaves the deployment environment. A single A100 running vLLM is sufficient; the cascade reaches \macrof{} above $0.94$ with no document, snippet, or prompt leaving the environment, and at roughly 40\% of the per-document LLM cost of zero-shot inference.

%% file: sections/08_industrial_validation.tex
\section{Financial-80: Production Deployment Results}
\label{sec:industrial}
\label{sec:fin80}

We validate \system{} on Financial-80, a private 80-category trade-finance document corpus drawn from a production financial-services pipeline. The setting is appreciably harder than Tobacco-3482: the category count is eight times larger, the distribution is more imbalanced, and many categories are near-synonyms (e.g., multiple invoice subtypes, multiple certificate variants) that require fine-grained disambiguation. Two documents from different near-synonym categories can share the majority of their visible text and differ only in a small set of boilerplate phrases or a single stamped clause, which is exactly the regime where retrieval alone is prone to voting for the wrong class.

\subsection{Dataset and class-count distribution}
\label{sec:fin80-dataset}

Financial-80 contains 32{,}911 documents across 80 categories. For the HITL-curve experiments, 648 seed documents initialise the index, 316 documents form the validation split and 1{,}714 the test split, and the remaining 30{,}233 unlabelled documents form the pool. The taxonomy is confidential; under the data-sharing agreement, per-class F1 tables and category names cannot be released. We report aggregate metrics only (overall \macrof{} and accuracy, head/medium/tail group F1, verifier call rate, and HITL queue rate).

\begin{figure}[t]
  \centering
  \includegraphics[width=1.0\linewidth]{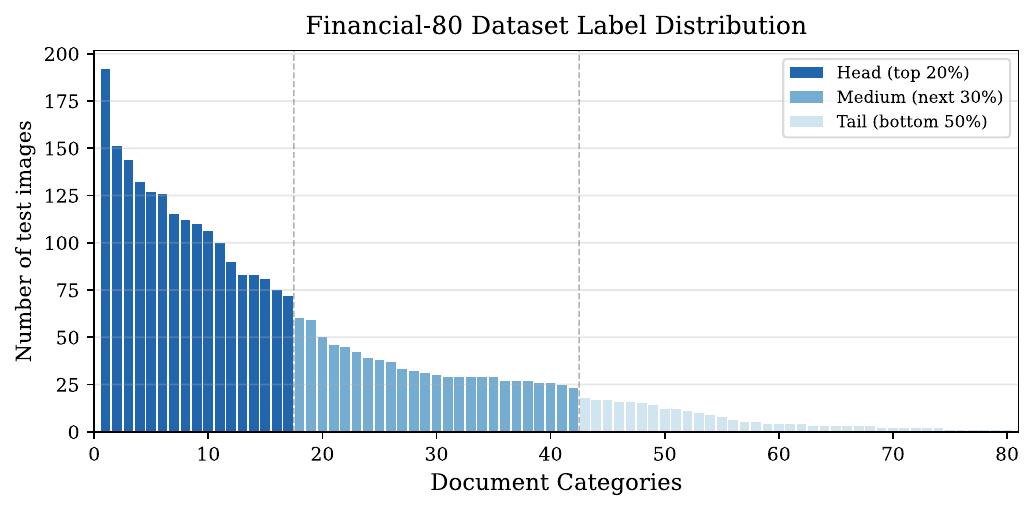}
  \caption{Class-count distribution of Financial-80 dataset.}
  \label{fig:fin80-dist}
\end{figure}

Figure~\ref{fig:fin80-dist} shows the class-count distribution: strongly imbalanced with a steeper slope than Tobacco-3482. Head (ranks 1--17, 17 classes), Medium (ranks 18--42, 25 classes), and Tail (ranks 43--80, 38 classes) groups account for approximately 60\%/30\%/10\% of volume.

\subsection{Static baselines}
\label{sec:fin80-block0}

Table~\ref{tab:fin80-block0} reports baselines (no HITL, seed-only index) on Financial-80 dataset. The same retrieval features and zero-shot LLM as Table~\ref{tab:block0} are used; the seed index is initialised with 648 seed documents (some long-tail categories have fewer than 10 samples).

\begin{table}[t]
\centering
\caption{Static baselines on Financial-80 dataset (seed-only index, 648 seed documents, $\sim$8 docs/class).}
\label{tab:fin80-block0}
\resizebox{\linewidth}{!}{%
\begin{tabular}{lrrrrr}
\toprule
Method & \macrof{} & Acc. & Head-F1 & Med-F1 & Tail-F1 \\
\midrule
BM25 only          & 0.351 & 0.484 & 0.517 & 0.435 & 0.307 \\
MiniLM only        & 0.332 & 0.476 & 0.463 & 0.505 & 0.242 \\
SigLIP2 only       & 0.391 & 0.554 & 0.681 & 0.514 & 0.316 \\
{RRF fusion}       & 0.502 & 0.663 & 0.749 & 0.623 & 0.438 \\
\midrule
\textbf{DS 0-shot}   & \underline{0.622} & \underline{0.824} & \underline{0.945} & \underline{0.740} & \underline{0.557} \\
\bottomrule
\end{tabular}%
}
\end{table}

Table~\ref{tab:fin80-block0} confirms that Financial-80 is harder than Tobacco-3482 (best baseline \macrof{} 0.622 vs.\ 0.774). RRF fusion improves over the best single retriever (\macrof{} 0.502 vs.\ 0.391), and the LLM head-bias is sharper on 80 categories: DS zero-shot achieves Head-F1 0.945 but Tail-F1 0.557, a 38.8-point Head--Tail gap (versus 26.8 on Tobacco). This gap matters operationally: a classifier that is excellent on the head categories but unreliable on the rest still forces a human to check most tail-category documents, which is exactly the workload \system{} is designed to shrink.

\subsection{\system{} cascade results}
\label{sec:fin80-hira}

\begin{figure}[t]
  \centering
  \includegraphics[width=1.0\linewidth]{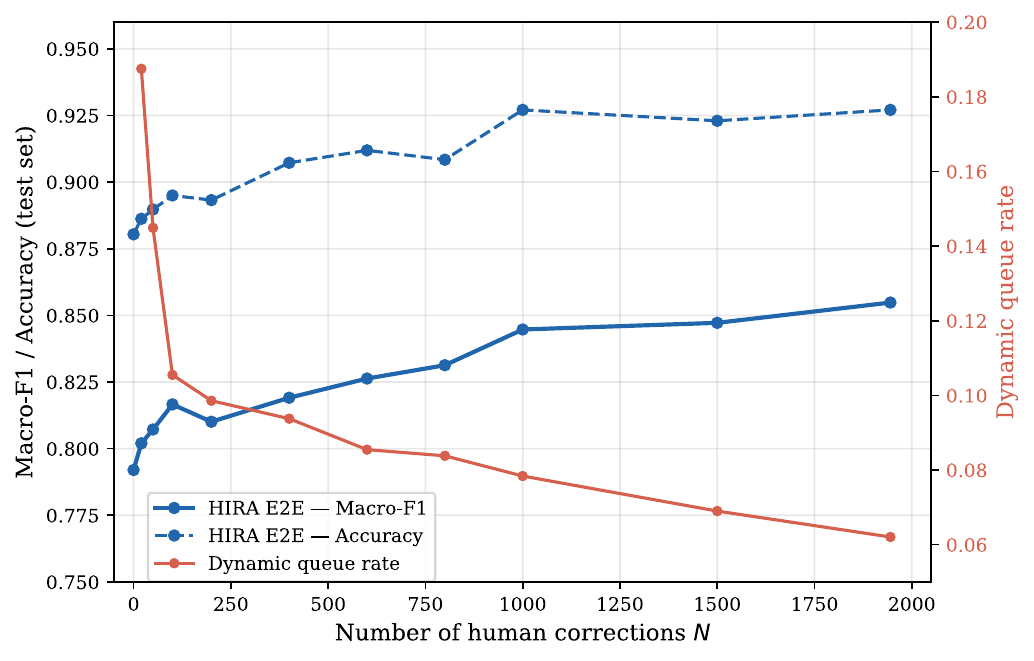}
  \caption{\system{} HITL improvement curve on Financial-80 (primary axis, green). Secondary axis (red): HITL dynamic queue rate.}
  \label{fig:fin80-hitl-curve}
\end{figure}

Figure~\ref{fig:fin80-hitl-curve} shows the HITL improvement trajectory on Financial-80 under the same V3 configuration ($\tauone=0.675$, $\tautwo=0.95$, $\tau_{\mathrm{index}}=10^4$) used on Tobacco-3482. V3 transferred without any dataset-specific recalibration. Table~\ref{tab:fin80-trajectory} reports the trajectory in tabular form.

\begin{table}[t]
\centering
\small
\caption{HIRA HITL trajectory on Financial-80 with end-to-end evaluation.}
\label{tab:fin80-trajectory}
\begin{tabular}{lrrrrr}
\toprule
$N$ & \macrof & Acc.\ & Head & Med & Tail \\
\midrule
DS 0-shot & 0.6218 & 0.8238 & 0.945 & 0.740 & 0.557 \\
\midrule
0    & 0.7920 & 0.8804 & 0.950 & 0.836 & 0.695 \\
200  & 0.8101 & 0.8932 & 0.965 & 0.848 & 0.687 \\
600  & 0.8263 & 0.9119 & 0.969 & 0.878 & 0.693 \\
1000 & 0.8447 & \underline{0.9271} & \underline{0.982} & 0.895 & 0.667 \\
\textbf{1945} & \underline{0.8548} & \underline{0.9271} & 0.976 & \underline{0.904} & \underline{0.701} \\
\bottomrule
\end{tabular}
\par\smallskip\footnotesize\raggedright
$N$ denotes the number of human corrections. By the final checkpoint, the system has scanned the full 30{,}233-document pool and requested labels for 1{,}945 documents.
\end{table}

\paragraph{HITL efficiency.}
After \system{} processes the entire 30{,}233-document stream, only 1{,}945 documents (6.4\%) are routed to human correction; at that final checkpoint \system{} reaches cascade-final \macrof{} 0.8548, an improvement of $+23.3$ points over the DS zero-shot baseline (0.6218). In absolute terms, 1{,}945 corrections took a single reviewer roughly six working days, well within the review team's existing capacity. The trajectory in Table~\ref{tab:fin80-trajectory} is still rising at the final evaluated checkpoint, suggesting further headroom as the index continues to mature under the more severe class imbalance.

\paragraph{EDRM and cascade contribution.}
Even at the modest checkpoint $N{=}1{,}945$, Tier-1-only retrieval (RRF fusion, no LLM verifier) reaches \macrof{} 0.7920 at $N{=}0$ while the full \system{} cascade reaches 0.8548 at $N{=}1{,}945$ --- a $+6.3$-point gain from the Tier-2 verifier and \edrm{} exemplar injection. Near-synonym categories (multiple invoice and certificate subtypes) dominate the confusion clusters here, which the distinguishing-term mechanism is designed to address.

\paragraph{Dynamic queue rate.}
At $N{=}1{,}945$, \textbf{61.6\%} of documents are accepted at Tier~1, \textbf{38.4\%} reach the LLM verifier at Tier~2, and \textbf{6.2\%} enter the HITL queue. This confirms that the verifier absorbs the majority of uncertain cases without human escalation. The secondary axis of Figure~\ref{fig:fin80-hitl-curve} shows that the dynamic queue rate, namely the fraction of incoming documents entering the HITL queue, decreases sharply as the system ingests more corrections, from approximately 18.8\% at the start to about 6.2\% at $N{=}1{,}945$. In practice, this mattered because review capacity --- not GPU capacity --- was the main operational bottleneck.

%% file: sections/10_limitations.tex
\section{Limitations}
\label{sec:limitations}

Most baselines are training-free; the DiT baselines are included only as limited-label supervised references on Tobacco-3482. We do not include a fully supervised LayoutLMv3 baseline on Financial-80 because collecting a sufficiently labelled training set is not feasible under the deployment constraints.

%% file: sections/11_conclusion.tex
\section{Conclusion}
\label{sec:conclusion}

We presented \system{}, a training-free, on-premises retrieval-augmented cascade for document classification in regulated industries, combining multimodal RRF retrieval, a locally hosted LLM verifier, and selective HITL correction stored as retrieval memory and confusion-graph updates. On Financial-80, \system{} reaches \macrof{} 0.8548 (from 0.6218) while routing only 6.4\% of the 30{,}233-document stream to human correction; on corrected Tobacco-3482, it reaches \macrof{} 0.9423, $+17.4$ points above LLM zero-shot, matching the fully labelled pool oracle with 24.8\% of the pool labelled. This combination of retrieval, a local verifier, and selective HITL was sufficient to ship without retraining the classifier.